\pdfoutput=1

\documentclass[11pt]{article}

\usepackage{acl}

\usepackage{times}
\usepackage{latexsym}

\usepackage[T1]{fontenc}

\usepackage[utf8]{inputenc}

\usepackage{microtype}

\usepackage{booktabs}
\usepackage{cleveref}
\crefformat{section}{\S#2#1#3}
\crefformat{subsection}{\S#2#1#3}
\crefformat{subsubsection}{\S#2#1#3}
\usepackage{graphicx}
\usepackage{CJKutf8}
\usepackage{multirow}
\usepackage{subfigure}
\usepackage{xcolor}

\newcommand{\ja}[1]{\begin{CJK}{UTF8}{min}#1\end{CJK}}
\newcommand{\minisection}[1]{\noindent{\bf {#1}.}}
\newcommand{\minisectionNoDot}[1]{\noindent{\bf {#1}}}
\usepackage{framed}

\newcommand{\Appendix}[1]{Appendix \ref{#1}}
\newcommand{\Table}[1]{Table \ref{#1}}
\newcommand{\Figure}[1]{Figure \ref{#1}}

\newcommand{\papertitle}{mLUKE: The Power of Entity Representations in Multilingual Pretrained Language Models}
\title{mLUKE: The Power of Entity Representations\\in Multilingual Pretrained Language Models}

\author{
    Ryokan Ri$^{1,2}$\thanks{\hspace{2mm}Work done as an intern at Studio Ousia.}\\
    {\small \texttt{ryo0123@ousia.jp}}
    \And
    Ikuya Yamada$^{1,3}$\\
    {\small \texttt{ikuya@ousia.jp}}
    \And
    Yoshimasa Tsuruoka$^{2}$\\
    {\small \texttt{tsuruoka@logos.t.u-tokyo.ac.jp}}
    \AND
    \begin{minipage}{\textwidth}
        \begin{center}
            \fontsize{11.5}{14}\selectfont
            \textnormal{$^1$Studio Ousia, Tokyo, Japan \\ $^2$The University of Tokyo, Tokyo, Japan \\ $^3$RIKEN AIP, Tokyo, Japan}\\ \end{center}
    \end{minipage}
}

\begin{document}
\maketitle
\begin{abstract}
Recent studies have shown that multilingual pretrained language models can be effectively improved with cross-lingual alignment information from Wikipedia entities.
However, existing methods only exploit entity information in pretraining and do not explicitly use entities in downstream tasks.
In this study, we explore the effectiveness of leveraging entity representations for downstream cross-lingual tasks.
We train a multilingual language model with 24 languages with entity representations and show
the model consistently outperforms word-based pretrained models in various cross-lingual transfer tasks.
We also analyze the model and the key insight is that incorporating entity representations into the input allows us to extract more language-agnostic features.
We also evaluate the model with a multilingual {\it cloze prompt} task with the mLAMA dataset.
We show that entity-based prompt elicits correct factual knowledge more likely than using only word representations.
Our source code and pretrained models are available at \url{https://github.com/studio-ousia/luke}.
\end{abstract}

\newcommand{\mask}{\texttt{[MASK]}}

\newcommand{\sentence}[1]{``{\it #1}''}
\newcommand{\word}[1]{{\it #1}}

\newcommand{\mluke}{mLUKE}
\newcommand{\mbert}{mBERT}
\newcommand{\xlmr}{XLM-R}
\newcommand{\extraTraining}{+ extra training}
\newcommand{\mlukeW}{mLUKE-W}
\newcommand{\mlukeE}{mLUKE-E}
\newcommand{\xlmk}{XLM-K}

\newcommand{\ba}{$\rm{_{base}}$}
\newcommand{\la}{$\rm{_{large}}$}

\newcommand{\X}{\texttt{[X]}}
\newcommand{\Y}{\texttt{[Y]}}
\newcommand{\mlukeEwithY}{\mlukeE{}\ba{} (\Y{})}
\newcommand{\mlukeEwithXY}{\mlukeE{}\ba{} (\X{} \& \Y{})}

\section{Introduction}
Pretrained language models have become crucial for achieving state-of-the-art performance in modern natural language processing.
In particular, multilingual language models \citep{NEURIPS2019_c04c19c2,conneau-etal-2020-unsupervised,Doddapaneni2021APO} have attracted considerable attention particularly due to their utility in cross-lingual transfer.

In zero-shot cross-lingual transfer, a pretrained encoder is fine-tuned in a single resource-rich language (typically English), and then evaluated on other languages never seen during fine-tuning.
A key to solving cross-lingual transfer tasks is to obtain representations that generalize well across languages.
Several studies aim to improve multilingual models with cross-lingual supervision such as bilingual word dictionaries \citep{conneau-etal-2020-emerging} or parallel sentences \citep{NEURIPS2019_c04c19c2}.

Another source of such information is the cross-lingual mappings of Wikipedia entities (articles).
Wikipedia entities are aligned across languages via inter-language links and the text contains numerous entity annotations (hyperlinks).
With these data, models can learn cross-lingual correspondence such as the words {\it Tokyo} (English) and \ja{{\it 東京}} (Japanese) refers to the same entity.
Wikipedia entity annotations have been shown to provide rich cross-lingual alignment information to improve multilingual language models \citep{Calixto2021naacl,XLM-K-2021-arxiv}.
However, previous studies only incorporate entity information through an auxiliary loss function during pretraining, and the models do not explicitly have entity representations used for downstream tasks.

In this study, we investigate the effectiveness of entity representations in multilingual language models.
Entity representations are known to enhance language models in mono-lingual settings \citep{Zhang2019,peters-knowbert,wang2019kepler,Xiong2020Pretrained,yamada-etal-2020-luke} presumably by introducing real-world knowledge.
We show that using entity representations facilitates cross-lingual transfer by providing language-independent features.
To this end, we present a multilingual extension of LUKE \citep{yamada-etal-2020-luke}.
The model is trained with the multilingual masked language modeling (MLM) task as well as the masked entity prediction (MEP) task with Wikipedia entity embeddings.

We investigate two ways of using the entity representations in cross-lingual transfer tasks:
(1) perform entity linking for the input text, and append the detected entity tokens to the input sequence. The entity tokens are expected to provide language-independent features to the model.
We evaluate this approach with cross-lingual question answering (QA) datasets: XQuAD \citep{artetxe-etal-2020-cross} and MLQA \citep{lewis-etal-2020-mlqa};
(2) use the entity \mask{} token from the MEP task as a language-independent feature extractor.
In the MEP task, word tokens in a mention span are associated with an entity \mask{} token, the contextualized representation of which is used to train the model to predict its original identity.
Here, we apply similar input formulations to tasks involving mention-span classification, relation extraction (RE) and named entity recognition (NER): the attribute of a mention or a pair of mentions is predicted using their contextualized entity \mask{} feature.
We evaluate this approach with the RELX \citep{koksal-ozgur-2020-relx} and CoNLL NER \citep{tjong-kim-sang-2002-introduction,TjongKimSang-DeMeulder:2003:CONLL} datasets.

The experimental results show that these entity-based approaches consistently outperform word-based baselines.
Our analysis reveals that entity representations provide more language-agnostic features to solve the downstream tasks.

We also explore solving a multilingual zero-shot {\it cloze prompt} task \citep{Liu2021PretrainPA} with the entity \mask{} token.
Recent studies have shown that we can address various downstream tasks by querying a language model for blanks in prompts \citep{petroni-etal-2019-language,cui-etal-2021-template}.
Typically, the answer tokens are predicted from the model's word-piece vocabulary but here we incorporate the prediction from the entity vocabulary queried by the entity \mask{} token.
We evaluate our approach with the mLAMA dataset \citep{kassner-etal-2021-multilingual} in various languages and show that using the entity \mask{} token reduces language bias and elicits correct factual knowledge more likely than using only the word \mask{} token.

\begin{figure*}[ht]

\begin{center}
  \includegraphics[width=14cm]{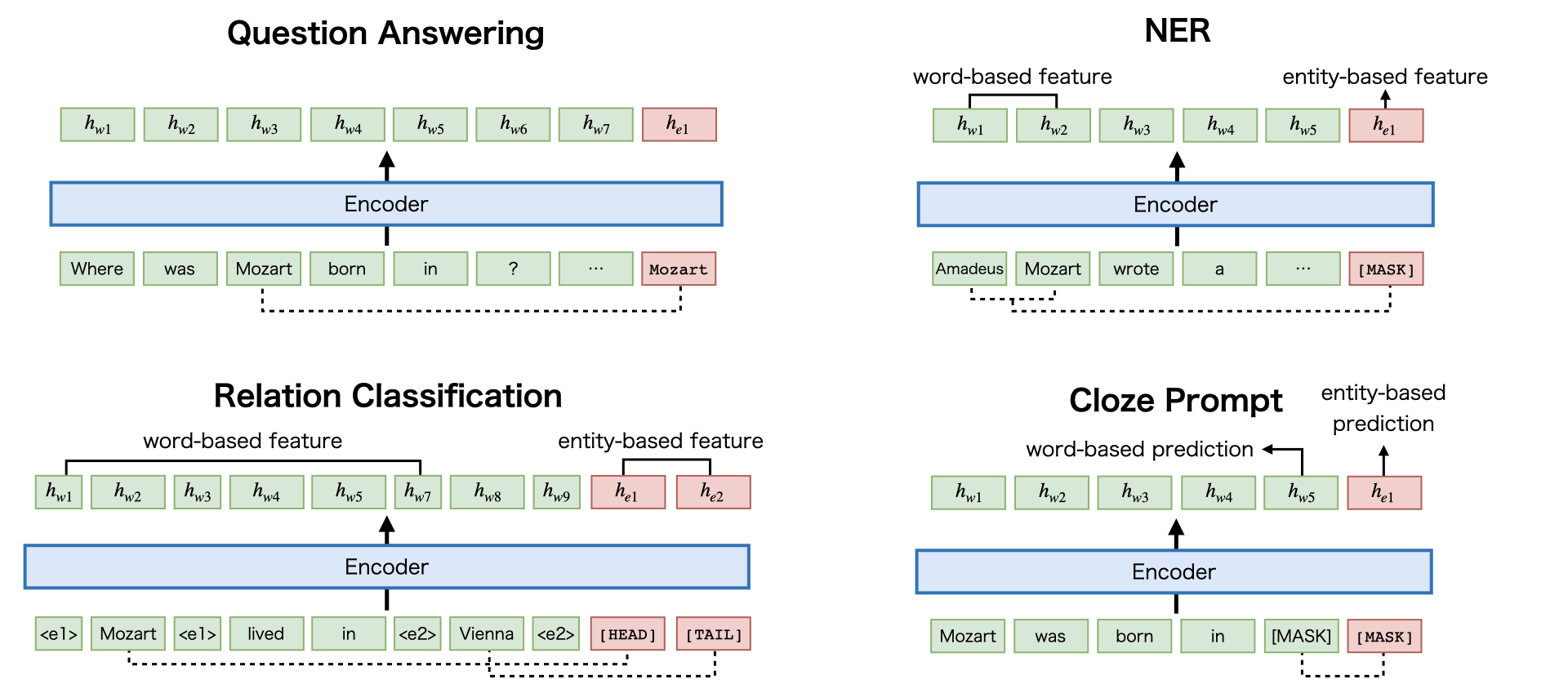}
  \caption{How to use entity representations in downstream tasks. The input entity embeddings are associated with their mentions (indicated by dotted lines) via positional embeddings.}
  \label{fig:downstream_models}
\end{center}
\vspace{-5mm}
\end{figure*}

\section{Multilingual Language Models with Entity Representations}

\subsection{Model: mulitlingual LUKE}
To evaluate the effectiveness of entity representations for cross-lingual downstream tasks, we introduce a new multilingual language model based on a bidirectional transformer encoder: Multilingual LUKE (\mluke{}), a multilingual extension of LUKE \citep{yamada-etal-2020-luke}.
The model is trained with the masked language modeling (MLM) task \citep{NIPS2017_7181} as well as the masked entity prediction (MEP) task. In MEP, some of the input entity tokens are randomly masked with the special entity \mask{} token, and the model is trained to predict the original entities.
Note that the entity \mask{} token is different from the word \mask{} token for MLM.

The model takes as input a tokenized text ($w_1, w_2, ..., w_m$) and the entities appearing in the text ($e_1, e_2, ..., e_n$), and compute the contextualized representation for each token ($\mathbf{h}_{w_1}, \mathbf{h}_{w_2}, ..., \mathbf{h}_{w_m}$ and $\mathbf{h}_{e_1}, \mathbf{h}_{e_2}, ..., \mathbf{h}_{e_n}$).
The word and entity tokens equally undergo self-attention computation ({\it i.e.}, no entity-aware self-attention in \citet{yamada-etal-2020-luke}) after embedding layers.

The word and entity embeddings are computed as the summation of the following three embeddings: token embeddings, type embeddings, and position embeddings \citep{devlin2018bert}.
The entity tokens are associated with the word tokens through position embeddings:
the position of an entity token is defined as the positions of its corresponding word tokens, and the entity position embeddings are summed over the positions.

\minisection{Model Configuration}
The model configurations of \mluke{} follow the {\it base} and {\it large} configurations of XLM-RoBERTa \citep{conneau-etal-2020-unsupervised}, a variant of BERT \citep{devlin2018bert} trained with CommonCrawl data from 100 languages.
Before pretraining, the parameters in common ({\it e.g.}, the weights of the transformer encoder and the word embeddings) are initialized using the checkpoint from the Transformers library.\footnote{\url{https://huggingface.co/transformers/}}

The size of the entity embeddings is set to 256 and they are projected to the size of the word embeddings before being fed into the encoder.

\subsection{Training Corpus: Wikipedia}
We use Wikipedia dumps in 24 languages (\Appendix{appendix:pretraining}) as the training data.
These languages are selected to cover reasonable numbers of languages that appear in downstream cross-lingual datasets.
We generate input sequences by splitting the content of each page into sequences of sentences comprising $\leq$ 512 words with their entity annotations ({\it i.e.}, hyperlinks).
During training, data are sampled from each language with $n_{i}$ items with the following multinomial distribution:

\begin{eqnarray}
  p_{i}=\frac{n_{i}^\alpha}{\sum_{k=1}^{N} n_{k}^\alpha},
\end{eqnarray}

\noindent
where $\alpha$ is a smoothing parameter and set to 0.7 following multilingual BERT.\footnote{\url{https://github.com/google-research/bert/blob/master/multilingual.md}}

\minisection{Entity Vocabulary}
Entities used in \mluke{} are defined as Wikipedia articles.
The articles from different languages are aligned through inter-language links\footnote{\url{https://en.wikipedia.org/wiki/Help:Interlanguage_links}. We build an inter-language database from the wikidatawiki dump from November 30, 2020.} and the aligned articles are treated as a single entity.
We include in the vocabulary the most frequent 1.2M entities in terms of the number of hyperlinks that appear across at least three languages to facilitate cross-lingual learning.

\minisection{Optimization}
We optimize the models with a batch size of 2048 for 1M steps in total using AdamW \citep{DBLP:conf/iclr/LoshchilovH19} with warmup and linear decay of the learning rate.
To stabilize training, we perform pretraining in two stages: (1) in the first 500K steps, we update only those parameters that are randomly initialized (e.g., entity embeddings); (2) we update all parameters in the remaining 500K steps.
The learning rate scheduler is reset at each training stage.
For further details on hyperparameters, see \Appendix{appendix:pretraining}.

\subsection{Baseline Models}

We compare the primary model that we investigate, multilingual LUKE used with entity representations ({\bf \mlukeE{}}), against several baselines pretrained models and an ablation model based on word representations:

\minisectionNoDot{\mbert{}} \citep{devlin2018bert} is one of the earliest multilingual language models. We provide these results as a reference.

\minisectionNoDot{\xlmr{}} \citep{conneau-etal-2020-unsupervised} is the model that \mluke{} is built on.
This result indicates how our additional pretraining step and entity representation impact the performance.
Since earlier studies \citep{Liu2019RoBERTaAR,lan2019albert} indicated longer pretraining would simply improve performance, we train another model based on \xlmr{}\ba{} with extra MLM pretraining following the same configuration of \mluke{}.

\minisectionNoDot{\mlukeW{}} is an ablation model of \mlukeE.
This model discards the entity embeddings learned during pretraining and only takes word tokens as input as with the other baseline models.
The results from this model indicate the effect of MEP only as an auxiliary task in pretraining, and the comparison with this model will highlight the effect of using entity representations for downstream tasks in \mlukeE{}.

The above models are fine-tuned with the same hyperparameter search space and computational budget as described in \Appendix{appendix:downstream}.

We also present the results of {\bf \xlmk{}} \citep{XLM-K-2021-arxiv} for ease of reference.
XLM-K is based on \xlmr{}\ba{} and trained with entity information from Wikipedia but does not use entity representations in downstream tasks.
Notice that their results are not strictly comparable to ours, because the pretraining and fine-tuning settings are different.

\begin{table*}[ht]
   \small
  \begin{tabular}{lccccccccccccc} \toprule
    XQuAD  &   en & es & de & el & ru & tr & ar & vi & th & zh & hi & avg. \\
  \midrule
  \mbert{} & 84.5 & 76.1 & 73.1 & 59.0 & 70.2 & 53.2 & 62.1 & 68.5 & 40.7 & 58.3 & 57.0 & 63.9 \\
  \xlmr{}\ba{} & 84.0 & 76.5 & 76.4 & 73.9 & 74.4 & 67.8 & 68.1 & 74.2 & 66.8 & 61.5 & 68.7 & 72.0 \\
  \extraTraining{} & 86.1 & 76.9 & 76.5 & 73.7 & 74.7 & 66.3 & 68.2 & 74.5 & {\bf 67.7} & 64.7 & 66.6 & 72.4 \\
  \mlukeW{}\ba{} & 85.7 & 78.0 & 77.4 & {\bf 74.7} & 75.7 & 68.3 & {\bf 71.7} & 75.9 & 67.1 & 65.1 & 69.9 & 73.6 \\
  \mlukeE{}\ba{} & {\bf 86.3} & {\bf 78.9} & {\bf 78.9} & 73.9 & {\bf 76.0} & {\bf 68.8} & 71.4 & {\bf 76.4} & 67.5 & {\bf 65.9} & {\bf 72.2} & {\bf 74.2} \\
  \midrule

  \xlmr{}\la{} &  88.5 & 82.4 & 82.0 & {\bf 81.4} & 81.2 & 75.5 & 75.9 & 80.7 & 72.3 & 67.6 & 77.2 & 78.6 \\
  \mlukeW{}\la{} & {\bf 89.0} & {\bf 83.1} & {\bf 82.4} & 81.3 & 81.3 & 75.3 & {\bf 77.9} & 81.2 & 75.1 & 71.5 & 77.3 & {\bf 79.6} \\
  \mlukeE{}\la{} & 88.6 & 83.0 & 81.7 & {\bf 81.4} & 80.8 & {\bf 75.8} & 77.7 & {\bf 81.9} & {\bf 75.4} & {\bf 71.9} & {\bf 77.5} & {\bf 79.6} \\
  
  \bottomrule
  \end{tabular}

  \begin{tabular}{lcccccccc|c} \toprule
    MLQA  &   en & es & de & ar & hi & vi & zh & avg. & G-XLT avg. \\ \midrule
  \mbert{} & 79.1 & 65.9 & 58.6 & 48.6 & 44.8 & 58.5 & 58.1 & 59.1 & 40.9 \\
  \xlmr{}\ba & 79.7 & 67.7 & 62.2 & 55.8 & 59.9 & 65.3 & 62.5 & 64.7 & 33.4 \\
  \extraTraining{} & {\bf 81.3} & 69.8 & 65.0 & 54.8 & 59.3 & 65.6 & 64.2 & 65.7 & 50.2 \\
  \mlukeW{}\ba{} & 81.3 & 69.7 & 65.4 & 60.4 & 63.2 & 68.3 & 66.1 & 67.8 & 54.0 \\
  \mlukeE{}\ba{} & 80.8 & {\bf 70.0} & {\bf 65.5} & {\bf 60.8} & {\bf 63.7} & {\bf 68.4} & {\bf 66.2} & {\bf 67.9} & {\bf 55.6} \\
  \midrule
  \xlmk{} \citep{XLM-K-2021-arxiv} & 80.8 & 69.2 & 63.8 & 60.0 & 65.3 & 70.1 & 63.8 & 67.7 & - \\
  \midrule
  \xlmr{}\la{} & 83.9 & {\bf 74.7} & 69.9 & 64.9 & 69.9 & 73.3 & 70.3 & 72.4 & 65.3 \\
  \mlukeW{}\la{} & 84.0 & 74.3 & 70.3 & {\bf 66.2} & 70.2 & 74.2 & 69.7 & 72.7 & 67.4 \\
  \mlukeE{}\la{} & {\bf 84.1} & 74.5 & {\bf 70.5} & {\bf 66.2} & {\bf 71.4} & {\bf 74.3} & {\bf 70.5} & {\bf 73.1} & {\bf 67.7} \\
  \bottomrule
  \end{tabular}

  \caption{F1 scores on the XQuAD and MLQA dataset in the cross-lingual transfer settings. The scores without reference are from the best model tuned with the English development data.}
  \label{table:results-qa}
  \vspace{-3mm}
\end{table*}
\normalsize

\vspace{-1mm}
\section{Adding Entities as Language-Agnostic Features in QA}
\label{sec:experiment-qa}
\vspace{-1mm}

We evaluate the approach of adding entity embeddings to the input of \mlukeE{} with cross-lingual extractive QA tasks.
The task is, given a question and a context passage, to extract the answer span from the context.
The entity embeddings provide language-agnostic features and thus should facilitate cross-lingual transfer learning.

\subsection{Main Experiments}

\minisection{Datasets}
We fine-tune the pretrained models with the SQuAD 1.1 dataset \citep{rajpurkar-etal-2016-squad}, and evaluate them with the two multilingual datasets: XQuAD \citep{artetxe-etal-2020-cross} and MLQA \citep{lewis-etal-2020-mlqa}.
XQuAD is created by translating a subset of the SQuAD development set while the source of MLQA is natural text in Wikipedia.
Besides multiple monolingual evaluation data splits, MLQA also offers data to evaluate generalized cross-lingual transfer (G-XLT), where the question and context texts are in different languages.

\minisection{Models}
All QA models used in this experiment follow \citet{devlin2018bert}.
The model takes the question and context word tokens as input and predicts a score for each span of the context word tokens. The span with the highest score is predicted as the answer to the question.

\mlukeE{} takes entity tokens as additional features in the input (\Figure{fig:downstream_models}) to enrich word representations.
The entities are automatically detected using a heuristic string matching based on the original Wikipedia article from which the dataset instance is created.
See Appendix \ref{sec:entity_detection} for more details.

\minisection{Results}
Table \ref{table:results-qa} summarizes the model's F1 scores for each language.
First, we discuss the {\it base} models.
On the effectiveness of entity representations, \mlukeE{}\ba{} performs better than its word-based counterpart \mlukeW{}\ba{} (0.6 average points improvement in the XQuAD average score, 0.1 points in MLQA) and \xlmk{} (0.2 points improvement in MLQA), which indicates the input entity tokens provide useful features to facilitate cross-lingual transfer.
The usefulness of entities is demonstrated especially in the MLQA's G-XLT setting (full results available in \Appendix{sec:full_result_mlqa});
\mlukeE{}\ba{} exhibits a substantial 1.6 point improvement in the G-XLT average score over \mlukeW{}\ba{}.
This suggests that entity representations are beneficial in a challenging situation where the model needs to capture language-agnostic semantics from text segments in different languages.

We also observe that \xlmr{}\ba{} benefits from extra training (0.4 points improvement in the average score on XQuAD and 2.1 points in MLQA).
The \mlukeW{}\ba{} model further improves the average score from \xlmr{}\ba{} with extra training (1.2 points improvement in XQuAD and 2.1 points in MLQA), showing the effectiveness of the MEP task for cross-lingual QA.

By comparing {\it large} models, we still observe substantial improvements from \xlmr{}\la{} to the \mluke{} models.
Also we can see that \mlukeE{}\la{} overall provides better results than \mlukeW{}\la{} (0.4 and 0.3 points improvements in the MLQA average and G-XLT scores; comparable scores in XQuAD), confirming the effectiveness of entity representations.

\subsection{Analysis}
\label{subsec:qa_analysis}

How do the entity representations help the model in cross-lingual transfer?
In the \mlukeE{} model, the input entity tokens annotate mention spans on which the model performs prediction.
We hypothesize that this allows the encoder to inject language-agnostic entity knowledge into span representations, which help better align representations across languages.
To support this hypothesis, we compare the degree of alignment between span representations before and after adding entity embeddings in the input, {\it i.e.}, \mlukeW{} and \mlukeE{}.

\minisection{Task}
We quantify the degree of alignment as performance on the contextualized word retrieval (CWR) task \citep{Cao2020MultilingualAO}. The task is, given a word within a sentence in the query language, to find the word with the same meaning in the context from a candidate pool in the target language.

\minisection{Dataset}
We use the MLQA dev set \citep{lewis-etal-2020-mlqa}.
As MLQA is constructed from parallel sentences mined from Wikipedia, some sentences and answer spans are aligned and thus the dataset can be easily adapted for the CWR task.
As the query and target word, we use the answer span\footnote{Answer spans are not necessarily a word, but here we generalize the task as span retrieval for our purpose.} annotated in the dataset, which is also parallel across the languages.
We use the English dataset as the query language and other languages as the target.
We discard query instances that do not have their parallel data in the target language.
The candidate pool is all answer spans in the target language data.

\minisection{Models}
We evaluate the \mlukeW{}\ba{} and \mlukeE{}\ba{} models without fine-tuning.
The retrieval is performed by ranking the cosine similarity of contextualized span representations, which is computed by mean-pooling the output word vectors in the span.

\minisection{Results}
Table \ref{table:squad_retrieval} shows the retrieval performance in terms of the mean reciprocal rank score.
We observe that the scores of \mlukeE{}\ba{} are higher than \mlukeW{}\ba{} across all the languages.
This demonstrates that adding entities improves the degree of alignment of span representations, which may explain the improvement of \mlukeE{} in the cross-lingual QA task.

\begin{table}[ht]
  \small
  \setlength\tabcolsep{3pt}
  \centering
  \begin{tabular}{lccccccc} \toprule
        &     ar      &     de     &     es     &     hi     &     vi     &     zh     &    avg.    \\ \midrule
\mlukeW{}\ba{}  &    55.6    &    66.1    &    68.4    &    60.4    &    69.7    &    56.1    &    62.7    \\
\mlukeE{}\ba{}  &    56.9    &    68.1    &    70.4    &    61.5    &    71.2    &    60.0    &    64.7    \\ \bottomrule
  \end{tabular}
  \caption{The mean reciprocal rank score of the CWR task with the MLQA dev set.}
  \label{table:squad_retrieval}
\end{table}

\begin{table*}[ht]
    \small
  \centering
\begin{tabular}{l|cccccl|ccccl} \toprule
                & \multicolumn{6}{c|}{RE} & \multicolumn{5}{c}{NER}         \\ \midrule
                & en   & de   & es   & fr   & tr   & avg. & en   & de   & nl   & es   & avg. \\ \midrule
\mbert{}       & 65.0 & 57.3 & 61.6 & 58.9 & 56.2 & 59.8 & 89.7 & 70.0 & 75.2 & 77.1 & 78.0 \\
\xlmr{}\ba{}        & 66.5 & 60.8 & 62.9 & 60.9 & 57.7 & 61.7 & 91.5 & 74.3 & 80.7 & {\bf 79.8} & 81.6 \\
\extraTraining{} & 67.0 & 61.3 & 62.9 & 64.3 & 61.9 & 63.5 & 91.8 & 75.7 & 80.3 & {\bf 79.8} & 81.9 \\
\mlukeW{}\ba{}      & 68.7 & 64.3 & {\bf 65.8} & 62.1 & 65.0 & 65.2 & 91.6 & 75.1 & 80.2 & 79.2 & 81.5 \\
\mlukeE{}\ba{}      & {\bf 69.3} & {\bf 64.5} & 65.2 & {\bf 64.7} & {\bf 68.7} & {\bf 66.5} & {\bf 93.6} & {\bf 77.2} & {\bf 81.8} & 77.7 & {\bf 82.6} \\
\midrule
XLM-K \citep{XLM-K-2021-arxiv} & - & - & - & - & - & - & 90.7 & 73.3 & 80.0 & 76.6 & 80.1 \\
\midrule
\xlmr{}\la{}        & 68.0 & 65.3 & 65.0 & 63.3 & 64.1 & 65.1 & 92.5 & 75.1 & 82.9 & 80.5 & 82.8 \\
\mlukeW{}\la{}      & 66.2 & 65.3 & {\bf 68.1} & {\bf 66.5} & {\bf 64.7} & 66.2 & 92.3 & 76.5 & 82.6 & 80.7 & 83.0 \\
\mlukeE{}\la{}      & {\bf 68.1} & {\bf 65.8} & 67.8 & 66.4 & 64.4 & {\bf 66.5} & {\bf 94.0} & {\bf 78.3} & {\bf 83.5} & {\bf 81.4} & {\bf 84.3} \\  
\bottomrule
\end{tabular}
\caption{F1 scores on relation extraction (RE) and named entity recognition (NER).}
\label{table:results-entity}
\vspace{-4mm}
\end{table*}

\section{The Entity MASK Token as Feature Extractor in RE and  NER}
\label{sec:experiment-rc-ner}
In this section, we evaluate the approach of using the entity \mask{} token to extract features from \mlukeE{} for two entity-related tasks: relation extraction and named entity recognition.

We formulate both tasks as the classification of mention spans.
The baseline models extract the feature of spans as the contextualized representations of word tokens, while \mlukeE{} extracts the feature as the contextualized representations of the special language-independent entity tokens associated with the mentions (Figure \ref{fig:downstream_models}).
We demonstrate that this approach consistently improves the performance in cross-lingual transfer.

\subsection{Relation Extraction}
\label{subsec:re}
Relation Extraction (RE) is a task to determine the correct relation between the two (head and tail) entities in a sentence.
Adding entity type features have been shown to be effective to cross-lingual transfer in RE \citep{subburathinam-etal-2019-cross,ahmad2020gatf}, but here we investigate an approach that does not require predefined entity types but utilize special entity embeddings learned in pretraining.

\minisection{Datasets}
We fine-tune the models with the English KBP-37 dataset \citep{Zhang2015RelationCV} and evaluate the models with the RELX dataset \citep{koksal-ozgur-2020-relx}, which is created by translating a subset of 502 sentences from KBP-37’s test set into four different languages.
Following \citet{koksal-ozgur-2020-relx}, we report the macro average of F1 scores of the 18 relations.

\minisection{Models}
In the input text, the head and tail entities are surrounded with special markers (\texttt{<ent>}, \texttt{<ent2>}).
The baseline models extract the feature vectors for the entities as the contextualized vector of the first marker followed by their mentions. The two entity features are concatenated and fed into a linear classifier to predict their relation.

For \mlukeE{}, we introduce two special entities, \texttt{[HEAD]} and \texttt{[TAIL]}, to represent the head and tail entities \citep{yamada-etal-2020-luke}. Their embeddings are initialized with the entity \mask{} embedding.
They are added to the input sequence being associated with the entity mentions in the input, and their contextualized representations are extracted as the feature vectors.
As with the word-based models, the features are concatenated and input to a linear classifier.

\subsection{Named Entity Recognition}
Named Entity Recognition (NER) is the task to detect entities in a sentence and classify their type.
We use the CoNLL-2003 English dataset \citep{TjongKimSang-DeMeulder:2003:CONLL} as the training data, and evaluate the models with the CoNLL-2003 German dataset and the CoNLL-2002 Spanish and Dutch dataset \citep{tjong-kim-sang-2002-introduction}.

\minisection{Models}
We adopt the model of \citet{Sohrab2018DeepRecognition} as the baseline model, which enumerates all possible spans in a sentence and classifies them into the target entity types or {\it non-entity} type. In this experiment, we enumerate spans with at most 16 tokens.
For the baseline models, the span features are computed as the concatenation of the word representations of the first and last tokens.
The span features are fed into a linear classifier to predict their entity type.

The input of \mlukeE{} contains the entity \mask{} tokens associated with all possible spans.
The span features are computed as the contextualized representations of the entity \mask{} tokens.
The features are input to a linear classifier as with the word-based models.

\subsection{Main Results}
The results are shown in Table \ref{table:results-entity}.
The \mlukeE{} models outperform their word-based counterparts \mlukeW{} in the average score in all the comparable settings (the {\it base} and {\it large} settings; the RE and NER tasks), which shows entity-based features are useful in cross-lingual tasks.
We also observe that \xlmr{}\ba{} benefits from extra training (1.8 average points improvement in RE and 0.3 points in NER), but \mlukeE{} still outperforms the results.

\begin{table}[ht]
  \centering
  \begin{tabular}{lcccc} \toprule
            & de   & es   & fr   & tr   \\ \midrule
  \mlukeW{}\ba{} & 0.71 & 0.74 & 0.74 & 0.84 \\
  \mlukeE{}\ba{} & 0.25 & 0.28 & 0.24 & 0.36 \\ \bottomrule
  \end{tabular}
  \caption{The modularity of word and entity features computed with the same mLUKE model. The data are from pairs of English and the other languages in the RELX dataset.}
  \label{table:modularity}
\end{table}

\subsection{Analysis}
The performance gain of \mlukeE{} over \mlukeW{} can be partly explained as the entity \mask{} token extracts better features for predicting entity attributes because it resembles how mLUKE is pretrained with the MEP task.
We hypothesize that there exists another factor for the improvement in cross-lingual performance: language neutrality of representations.

The entity \mask{} token is shared across languages and their contextualized representations may be less affected by the difference of input languages, resulting in features that generalize well for cross-lingual transfer.
To find out if the entity-based features are actually more language-independent than word-based features, we evaluate the {\it modularity} \citep{fujinuma-etal-2019-resource} of the features extracted for the RELX dataset.

Modularity is computed for the $k$-nearest neighbor graph of embeddings and measures the degree to which embeddings tend to form clusters within the same language.
We refer readers to \citet{fujinuma-etal-2019-resource} for how to compute the metric.
Note that the maximum value of modularity is 1, and 0 means the embeddings are completely randomly distributed regardless of language.

\begin{table*}[ht]
  \small
  \centering
  \begin{tabular}{l|cccccccccc} \toprule
        &   ar & en & fi & fr & id & ja & ru & vi & zh & avg. \\ \midrule
\mbert{} & 17.1 & 36.8 & 24.0 & 24.3 & 42.9 & 14.3 & 19.5 & 39.4 & 26.2 & 27.2 \\
\xlmr{}\ba & 14.2 & 27.2 & 16.2 & 14.9 & 28.2 & 11.9 & 11.7 & 25.1 & 17.6 & 18.5 \\
\extraTraining{} & 21.2 & 35.0 & 23.0 & 22.2 & 46.8 & 19.6 & 17.5 & 34.4 & 30.7 & 27.8 \\
\mlukeW{}\ba & 22.3 & 31.3 & 18.4 & 19.6 & 46.7 & 18.4 & 16.7 & 31.9 & 29.3 & 26.1 \\
\mlukeEwithY{} & 27.8 & 37.5 & 30.4 & 28.4 & 44.2 & 28.9 & 25.8 & 42.1 & 33.4 & 33.2 \\
\mlukeEwithXY{} & {\bf 42.4} & {\bf 47.5} & {\bf 44.2} & {\bf 35.9} & {\bf 56.2} & {\bf 40.3} & {\bf 35.5} & {\bf 55.2} & {\bf 46.7} & {\bf 44.9} \\
\bottomrule
  \end{tabular}
      \vspace{-1mm}
  \caption{The top-1 accuracies from 9 languages from the mLAMA dataset.}
  \label{table:mlama}
    \vspace{-1mm}
\end{table*}

We compare the modularity of the word features from \mlukeW{}\ba{} and entity features from \mlukeE{}\ba{} before fine-tuning.
Note that the features here are concatenated vectors of head and tail features.
Table \ref{table:modularity} shows that the modularity of \mlukeE{}\ba{} is much lower than \mlukeW{}\ba{}, demonstrating that entity-based features are more language-neutral.
However, with entity-based features, the modularities are still greater than zero.
In particular, the modularity computed with Turkish, which is the most distant language from English here, is significantly higher than the others, indicating that the contextualized entity-based features are still somewhat language-dependent.

\section{Cloze Prompt Task with Entity Representations}

In this section, we show that using the entity representations is effective in a cloze prompt task \citep{Liu2021PretrainPA} with the mLAMA dataset \citep{kassner-etal-2021-multilingual}.
The task is, given a cloze template such as \sentence{\X{} was born in \Y{}} with \X{} filled with an entity ({\it e.g.}, \word{Mozart}), to predict a correct entity in \Y{} ({\it e.g.,} \word{Austria}).
We adopt the typed querying setting \citep{kassner-etal-2021-multilingual}, where a template has a set of candidate answer entities and the prediction becomes the one with the highest score assigned by the language model.

\minisection{Model}
As in \citet{kassner-etal-2021-multilingual}, the word-based baseline models compute the candidate score as the log-probability from the MLM classifier.
When a candidate entity in \Y{} is tokenized into multiple tokens, the same number of the word \mask{} tokens are placed in the input sequence, and the score is computed by taking the average of the log-probabilities for its individual tokens.

On the other hand, \mlukeE{} computes the log-probability of the candidate entity in \Y{} with the entity \mask{} token.
Each candidate entity is associated with an entity in mLUKE's entity vocabulary via string matching.
The input sequence has the entity \mask{} token associated with the word \mask{} tokens in \Y{}, and the candidate score is computed as the log-probability from the MEP classifier.
We also try additionally appending the entity token of \X{} to the input sequence if the entity is found in the vocabulary.

To accurately measure the difference between word-based and entity-based prediction, we restrict the candidate entities to the ones found in the entity vocabulary and exclude the questions if their answers are not included in the candidates (results with full candidates and questions in the dataset are in \Appendix{sec:full_mlama}).

\minisection{Results}
We experiment in total with 16 languages which are available both in the mLAMA dataset and the \mluke{}'s entity vocabulary.
Here we only present the top-1 accuracy results from 9 languages on \Table{table:mlama}, as we can make similar observations with the other languages.

We observe that \xlmr{}\ba{} performs notably worse than \mbert{} as mentioned in \citet{kassner-etal-2021-multilingual}.
However, with extra training with the Wikipedia corpus, \xlmr{}\ba{} shows a significant 9.3 points improvement in the average score and outperforms \mbert{} (27.8 vs. 27.2).
We conjecture that this shows the importance of the training corpus for this task.
The original \xlmr{} is only trained with the CommonCrawl corpus \citep{conneau-etal-2020-unsupervised}, text scraped from a wide variety of web pages, while \mbert{} and \xlmr{} + training are trained on Wikipedia.
The performance gaps indicate that Wikipedia is particularly useful for the model to learn factual knowledge.

The \mlukeW{}\ba{} model lags behind \xlmr{}\ba{} \extraTraining{} by 1.7 average points but we can see 5.4 points improvement from \xlmr{}\ba{} \extraTraining{} to \mlukeEwithY{}, indicating entity representations are more suitable to elicit correct factual knowledge from \mluke{} than word representations.
Adding the entity corresponding to \X{} to the input (\mlukeEwithXY{}) further pushes the performance by 11.7 points to 44.9 \%, which further demonstrates the effectiveness of entity representations.

\begin{table*}[ht]
  \small
  \centering
  \begin{tabular}{lrrr} \toprule
                      & \multicolumn{1}{c}{en}  & \multicolumn{1}{c}{ja} & \multicolumn{1}{c}{fr}                   \\ \midrule
  \mbert{}            & The Bahamas, 41\% (355/870) & Japan, 82\% (361/439) & Pays-Bas, 71\% (632/895)  \\
  \xlmr{}\ba             & London, 78\% (664/850)      & Japan, 99\% (437/440) & Allemagne, 96\% (877/916)  \\
  \extraTraining{}    & Australia, 27\% (247/899)   & Japan, 99\% (437/442) & Allemagne, 93\% (854/917) \\
  \mlukeW{}\ba           & Germany, 22\% (198/895)     & Japan, 97\% (428/442) & Allemagne, 99\% (906/918) \\
  \mlukeEwithY{}      & London, 37\% (310/846)      & Japan, 56\% (241/430) & Suède, 40\% (362/908)     \\
  \mlukeEwithXY{}     & London, 27\% (213/797)      & Japan, 44\% (176/401) & Suède, 30\% (266/895) \\ \bottomrule
  \end{tabular}
  \caption{The top incorrect predictions in three languages for the template \sentence{\X{} was founded in \Y{}.} for each model. The predictions in the original language are translated into English.}
  \label{table:top-k_false_positive_prediction}
  \normalsize
  \vspace{-3mm}
\end{table*}

\minisection{Analysis of Language Bias}
\citet{kassner-etal-2021-multilingual} notes that the prediction of \mbert{} is biased by the input language.
For example, when queried in Italian ({\it e.g.}, \sentence{\X{} e stato creato in \mask{}.}), the model tends to predict entities that often appear in Italian text ({\it e.g.,} \word{Italy}) for any question to answer location.
We expect that using entity representations would reduce language bias because entities are shared among languages and less affected by the frequency in the language of  questions.

We qualitatively assess the degree of language bias in the models looking at their incorrect predictions.
We show the top incorrect prediction for the template \sentence{\X{} was founded in \Y{}.} for each model in \Table{table:top-k_false_positive_prediction}, together with {\it the top-$1$ incorrect ratio}, that is, the ratio of the number of the most common incorrect prediction to the total false predictions, which indicates how much the false predictions are dominated by few frequent entities.

The examples show that the different models exhibit bias towards different entities as in English and French, although in Japanese the model consistently tends to predict \word{Japan}.
Looking at the degree of language bias, \mlukeEwithXY{} exhibits lower top-$1$ incorrect ratios overall (27\% in fr, 44\% in ja, and 30\% in fr), which indicates using entity representations reduces language bias.
However, lower language bias does not necessarily mean better performance: in French (fr), \mlukeEwithXY{} gives
a lower top-$1$ incorrect ratio than \mbert{} (30\% vs. 71\%) but their numbers of total false predictions are the same (895).
Language bias is only one of several factors in the performance bottleneck.

\section{Related Work}

\subsection{Multilingual Pretrained Language Models}
Multilingual pretrained language models have recently seen a surge of interest due to their effectiveness in cross-lingual transfer learning \citep{NEURIPS2019_c04c19c2,liu-etal-2020-multilingual}.
A straightforward way to train such models is multilingual masked language modeling (mMLM) \citep{devlin2018bert,conneau-etal-2020-unsupervised}, i.e., training a single model with a collection of monolingual corpora in multiple languages.
Although models trained with mMLM exhibit a strong cross-lingual ability without any cross-lingual supervision \citep{K-mBERT-ICLR-2020,conneau-etal-2020-emerging}, several studies aim to develop better multilingual models with explicit cross-lingual supervision such as bilingual word dictionaries \citep{conneau-etal-2020-emerging} or parallel sentences \citep{NEURIPS2019_c04c19c2}.
In this study, we build a multilingual pretrained language model on the basis of XLM-RoBERTa \citep{conneau-etal-2020-unsupervised}, trained with mMLM as well as the masked entity prediction (MEP) \citep{yamada-etal-2020-luke} with entity representations.

\subsection{Pretrained Language Models with Entity Knowledge}
Language models trained with a large corpus contain knowledge about real-world entities, which is useful for entity-related downstream tasks such as relation classification, named entity recognition, and question answering.
Previous studies have shown that we can improve language models for such tasks by incorporating entity information into the model \citep{Zhang2019,peters-knowbert,wang2019kepler,Xiong2020Pretrained,fevry-etal-2020-entities,yamada-etal-2020-luke}.

When incorporated into multilingual language models, entity information can bring another benefit: entities may serve as anchors for the model to align representations across languages.
Multilingual knowledge bases such as Wikipedia often offer mappings between different surface forms across languages for the same entity.
\citet{Calixto2021naacl} fine-tuned the top two layers of multilingual BERT by predicting language-agnostic entity ID from hyperlinks in Wikipedia articles.
As our concurrent work, \citet{XLM-K-2021-arxiv} trained a model based on XLM-RoBERTa with an entity prediction task along with an object entailment prediction task.
While the previous studies focus on improving cross-lingual language representations by pretraining with entity information, our work investigates a multilingual model not only pretrained with entities but also explicitly having entity representations and how to extract better features from such model.

\section{Conclusion}
We investigated the effectiveness of entity representations in multilingual language models.
Our pretrained model, mLUKE, not only exhibits strong empirical results with the word inputs (\mlukeW{}) but also shows even better performance with the entity representations (\mlukeE{}) in cross-lingual transfer tasks.
We also show that a cloze-prompt-style fact completion task can effectively be solved with the query and answer space in the entity vocabulary.
Our results suggest a promising direction to pursue further on how to leverage entity representations in multilingual tasks.
Also, in the current model, entities are represented as individual vectors, which may incur a large memory footprint in practice.
One can investigate an efficient way of having entity representations.

 \bibliography{references/references}
 \bibliographystyle{acl_natbib}

\appendix
\onecolumn
\section*{Appendix for ``\papertitle{}''}

\section{Details of Pretraining}
\label{appendix:pretraining}

\minisection{Dataset}
We download the Wikipedia dumps from December 1st, 2020.
We show the 24 languages included in the dataset on Table \ref{table:data_stat}, along with the data size and the number of entities in the vocabulary.

\begin{table}[h]
  \centering
  \begin{tabular}{crr|crr} \toprule
  Language Code & \multicolumn{1}{c}{Size} & \multicolumn{1}{c|}{\# entities in vocab} & Language Code & \multicolumn{1}{c}{Size} & \multicolumn{1}{c}{\# entities in vocab} \\ \midrule
  ar & 851M & 427,460 &  ko & 537M & 378,399 \\
  bn & 117M & 62,595  &  nl & 1.1G & 483,277 \\
  de & 3.5G & 540,347 &  pl & 1.3G & 489,109 \\
  el & 315M & 135,277 &  pt & 1.0G & 537,028 \\
  en & 6.9G & 613,718 &  ru & 2.5G & 529,171 \\
  es & 2.1G & 587,525 &  sv & 1.1G & 390,313 \\
  fi & 480M & 300,333 &  sw &  27M & 30,129 \\
  fr & 3.1G & 630,355 &  te &  66M & 14,368 \\
  hi &  90M & 54,038  &  th & 153M & 100,231 \\
  id & 327M & 217,758 &  tr & 326M & 297,280 \\
  it & 1.9G & 590,147 &  vi & 516M & 263,424 \\
  ja & 2.3G & 369,470 &  zh & 955M & 332,970 \\ \midrule
     &      &        &  Total & 31.4G & 8,374,722 \\ \bottomrule
  \end{tabular}
  \caption{Training Data Statistics: the size of training data, and the number of entities found in the 1.2M entity vocabulary.}
  \label{table:data_stat}
\end{table}

\minisection{Optimization}
We optimize the \mluke{} models for 1M steps in total using AdamW \citep{DBLP:conf/iclr/LoshchilovH19} with learning rate warmup and linear decay of the learning rate.
The pretraining consists of two stages: (1) in the first 500K steps, we update only those parameters that are randomly initialized (e.g., entity embeddings); (2) we update all parameters in the remaining 500K steps.
The learning rate scheduler is reset at each training stage.
The detailed hyper-parameters are shown in Table \ref{tb:pretraining-config}.

\begin{table}[h]
    \centering

    \begin{tabular}{lc|lc}
        \toprule
        Maximum word length                   & 512    & Mask probability for entities         & 15\%   \\
        Batch size                            & 2048   & The size of word token embeddings & 768   \\
        Peak learning rate                    & 1e-4   & The size of entity token embeddings         & 256 \\ 
        Peak learning rate (first 500K steps) & 5e-4   & Dropout                               & 0.1   \\
        Learning rate decay                   & linear & Weight decay                          & 0.01   \\
        Warmup steps                          & 2500   & Adam $\beta_1$                        & 0.9    \\
        Mask probability for words            & 15\%   & Adam $\beta_2$                        & 0.999  \\
        Random-word probability for words     & 10\%   & Adam $\epsilon$                       & 1e-6   \\
        Unmasked probability for words        & 10\%   & Gradient clipping                     & none \\
        \bottomrule
    \end{tabular}

    \caption{Hyper-parameters used to pretrain \mluke{}.}
    \label{tb:pretraining-config}
\end{table}

\minisection{Computing Infrastructure}
We run the pretraining on NVIDIA's PyTorch Docker container 19.02 hosted on a server with two Intel Xeon Platinum 8168 CPUs and 16 NVIDIA Tesla V100 GPUs.
The training takes approximately 2 months.

\newpage

\section{Details of Downstream Experiments}
\label{appendix:downstream}

\minisection{Hyperparameter Search}
For each downstream task, we perform hyperparameter searching for all the models with the same computational budget to ensure a fair comparison.
For each task, we use the final evaluation metric on the validation split of the training English corpus as the validation score.
The models are optimized with the AdamW optimizer \citep{DBLP:conf/iclr/LoshchilovH19} with the weight decay term set to 0.01 and a linear warmup scheduler. The learning rate is linearly increased to a specified value in the first 6 \% of training steps, and then gradually decreased to zero towards the end.
\Table{table:hyperparameters} summarizes the task-specific hyperparameter search spaces.

\begin{table}[h]
 \centering
 \begin{tabular}{lccc} \toprule
                     & \begin{tabular}[c]{@{}c@{}}QA\\ (SQuAD)\end{tabular} & \begin{tabular}[c]{@{}c@{}}Relation Classification\\ (KBP37)\end{tabular} & \begin{tabular}[c]{@{}c@{}}NER\\ (CoNLL 2003)\end{tabular} \\ \midrule
  Learning rate      & 2e-5                                                 & 2e-5                                                                      & 2e-5                                                       \\
  Batch size         & \{16, 32\}                                           & \{4, 8, 16\}                                                              & \{4, 8, 16\}                                               \\
  Epochs             & 2                                                    & 5                                                                         & 5                                                          \\
  \# of random seeds & 3                                                    & 3                                                                         & 3 \\
  Validation metric & F1 & F1 & F1 \\
  \bottomrule
  
  \end{tabular}
  \caption{The hyperparameters search spaces and other details of downstream experiments.}
  \label{table:hyperparameters}
\end{table}

\minisection{Computing Infrastructure}
We run the fine-tuning on a server with a Intel(R) Core(TM) i7-6950X CPU and 4 NVIDIA GeForce RTX 3090 GPUs.

\section{Detecting Entities in the QA datasets}
\label{sec:entity_detection}
For each question–passage pair in the QA datasets, we first create a mapping from the entity mention strings (e.g., “U.S.”) to their referent Wikipedia entities (e.g., United States) using the entity hyperlinks on the source Wikipedia page of the passage. We then perform simple string matching to extract all entity names in the question and the passage and treat all matched entity names as entity annotations for their referent entities. We ignore an entity name if the name refers to multiple entities on the page.
Further, to reduce noise, we also exclude an entity name if its link probability, the probability that the name appears as a hyperlink in Wikipedia, is lower than 1\%.

The XQuAD datasets are created by translating English Wikipedia articles into target languages.
For each translated article, we create the mention-entity mapping from the source English article by the following procedure: for all the entities found in the source article, we find the corresponding entity in the target language through inter-language links, and then collect its possible mention strings ({\it i.e., } hyperlinks to the entity) from a Wikipedia dump of the target language; the entity and the collected mention strings form the mention-entity mapping for the translated article.


\newpage
\section{The Model Size}
\label{appendix:model_size}

\begin{table}[h]
\centering
\begin{tabular}{lccccc}  \toprule
        & \# of layers & hidden size & \multicolumn{1}{l}{\# of heads} & vocabulary size & \# of parameters \\
        \midrule
\mbert{}   & 12           & 768         & 12                              & 120K            & 177M             \\
\xlmr{}\ba   & 12           & 768         & 8                               & 250K            & 278M             \\
\mlukeE{}\ba & 12           & 768         & 8                               & 250K            & 585M  \\
\xlmr{}\la   & 24           & 1024         & 16                               & 250K            & 559M             \\
\mlukeE{}\la & 24           & 1024         & 16                               & 250K            & 867M  \\
\bottomrule
\end{tabular}

  \caption{The model sizes of the pretrained models.}
  \label{table:model_size}
\end{table}
\section{Ablation Study of Entity Embeddings}
\label{appendix:entity-ablation}

\newcommand{\ablation}{- ablation}

In Section \ref{sec:experiment-qa} and \ref{sec:experiment-rc-ner}, we have shown that using entity representations in \mluke{} improves the cross-lingual transfer performance in QA, RE, and NER.
Here we conduct an additional ablation study to investigate whether the learned entity embeddings are crucial to the success of our approach.
We train an ablated model of \mlukeE{} whose entity embeddings are re-initialized randomly before fine-tuning (\ablation{}).
\Table{table:results-qa-random-ablation} and \Table{table:results-entity-random-ablation} show that the ablated model performs significantly worse than the full model (\mlukeE{}), indicating that using pretrained entity embeddings is crucial rather than applying our approach during fine-tuning in an ad-hoc manner without entity-aware pretraining. 

\begin{table*}[ht]
  \begin{tabular}{lccccccccccccc} \toprule
      XQuAD  &   en & es & de & el & ru & tr & ar & vi & th & zh & hi & avg. \\ \midrule
\mlukeE{} & 86.3 & 78.9 & 78.9 & 73.9 & 76.0 & 68.8 & 71.4 & 76.4 & 67.5 & 65.9 & 72.2 & 74.2 \\ 
\ablation{} & 84.3 & 76.8 & 76.4 & 71.9 & 74.3 & 67.4 & 70.2 & 75.3 & 67.1 & 64.4 & 68.4 & 72.4 \\
  \bottomrule
  \end{tabular}

  \begin{tabular}{lcccccccc|c} \toprule
    MLQA  &   en & es & de & ar & hi & vi & zh & avg. & G-XLT avg. \\ \midrule
\mlukeE{}\ba & 80.8 & 70.0 & 65.5 & 60.8 & 63.7 & 68.4 & 66.2 & 67.9 & 55.6 \\
\ablation{} & 80.3 & 69.4 & 64.5 & 59.1 & 59.2 & 66.5 & 63.6 & 66.1 & 50.7 \\
  \bottomrule
  \end{tabular}

  \caption{F1 scores on the XQuAD and MLQA datasets in the cross-lingual transfer settings.}
  \label{table:results-qa-random-ablation}
\end{table*}
\normalsize

\begin{table*}[ht]
  \centering
\begin{tabular}{l|cccccl|ccccl} \toprule
                & \multicolumn{6}{c|}{RE} & \multicolumn{5}{c}{NER}         \\ \midrule
                & en   & de   & es   & fr   & tr   & avg. & en   & de   & du   & es   & avg. \\ 
\mlukeE{}\ba      & 69.3 & 64.5 & 65.2 & 64.7 & 68.7 & 66.5 & 93.6 & 77.2 & 81.8 & 77.7 & 82.6 \\
\ablation{} & 62.5 & 59.3 & 60.7 & 61.0 & 60.5 & 50.8 & 93.0 & 76.3 & 80.8 & 76.1 & 81.6 \\
\bottomrule
\end{tabular}
\caption{F1 scores on relation extraction (RE) and named entity recognition (NER).}
\label{table:results-entity-random-ablation}
\end{table*}

\newpage

\section{Full Results of MLQA}
\label{sec:full_result_mlqa}

\begin{table}[h!]
  \small
  \center
  \begin{minipage}[t]{.45\textwidth}
    \begin{center}
  \begin{tabular}{cccccccc} \toprule
    c/q & en & es & de & ar & hi & vi & zh \\ \midrule
    en & 79.1 & 65.4 & 63.4 & 37.9 & 29.7 & 47.1 & 43.2 \\
    es & 67.7 & 65.9 & 58.2 & 38.2 & 24.4 & 43.6 & 39.5 \\
    de & 61.7 & 55.9 & 58.6 & 32.3 & 29.7 & 38.4 & 36.8 \\
    ar & 49.9 & 43.2 & 44.6 & 48.6 & 23.4 & 29.4 & 27.1 \\
    hi & 47.0 & 37.8 & 39.1 & 26.2 & 44.8 & 28.0 & 23.0 \\
    vi & 59.9 & 49.4 & 48.6 & 26.7 & 25.6 & 58.5 & 40.7 \\
    zh & 55.3 & 44.2 & 45.3 & 28.3 & 22.7 & 38.7 & 58.1 \\ \bottomrule
  \end{tabular}
  \end{center}
  \caption{MLQA full results of \mbert{}}.
  \label{table:mlqa-mBERT}
  \end{minipage}
  \hspace{5mm}
  \begin{minipage}[t]{.45\textwidth}
    \begin{center}
      \begin{tabular}{cccccccc} \toprule

      c/q & en & es & de & ar & hi & vi & zh \\ \midrule
      en & 79.6 & 52.3 & 59.6 & 30.8 & 43.2 & 40.0 & 36.0 \\
      es & 67.0 & 67.7 & 52.0 & 25.2 & 31.8 & 32.9 & 31.5 \\
      de & 59.5 & 41.7 & 62.1 & 22.2 & 27.8 & 29.2 & 29.5 \\
      ar & 49.6 & 23.2 & 30.9 & 55.8 & 10.6 & 11.6 & 10.3 \\
      hi & 58.5 & 34.6 & 42.3 & 17.8 & 59.8 & 22.4 & 23.0 \\
      vi & 61.1 & 28.1 & 39.5 & 17.0 & 27.5 & 65.2 & 26.5 \\
      zh & 55.2 & 22.7 & 28.1 & 9.26 & 21.1 & 17.5 & 62.4 \\ \bottomrule

      \end{tabular}
    \end{center}
    \caption{MLQA full results of \xlmr{}\ba{}}
    \label{table:mlqa-xlmr-base}
  \end{minipage}
\end{table}

\begin{table}[h!]
  \small
  \center
  \begin{minipage}[t]{.45\textwidth}
    \begin{center}
      \begin{tabular}{cccccccc} \toprule

      c/q & en & es & de & ar & hi & vi & zh \\ \midrule
      en & 81.3 & 71.2 & 70.1 & 40.6 & 52.3 & 54.8 & 48.2 \\
      es & 70.6 & 69.8 & 66.2 & 43.3 & 47.9 & 52.8 & 49.0 \\
      de & 64.4 & 60.4 & 64.9 & 36.8 & 42.3 & 44.3 & 42.9 \\
      ar & 59.3 & 52.3 & 52.2 & 54.8 & 30.3 & 37.1 & 31.5 \\
      hi & 65.0 & 56.5 & 56.8 & 33.8 & 59.3 & 43.0 & 39.9 \\
      vi & 67.0 & 57.1 & 58.2 & 31.7 & 43.8 & 65.5 & 44.0 \\
      zh & 62.4 & 53.7 & 54.2 & 33.3 & 40.2 & 44.8 & 64.2 \\ \bottomrule

      \end{tabular}
    \end{center}
    \caption{MLQA full results of \xlmr{}\ba{} + training}
    \label{table:mlqa-xlmr-base-train}
  \end{minipage}
  \hspace{5mm}
  \begin{minipage}[t]{.45\textwidth}
    \begin{center}
      \begin{tabular}{cccccccc} \toprule
        c/q & en & es & de & ar & hi & vi & zh \\ \midrule
        en & 81.2 & 69.5 & 69.1 & 53.6 & 60.8 & 60.4 & 58.4 \\
        es & 70.3 & 69.6 & 65.5 & 52.1 & 52.9 & 56.1 & 56.4 \\
        de & 64.7 & 59.8 & 65.3 & 45.4 & 48.9 & 49.9 & 49.3 \\
        ar & 60.4 & 52.3 & 54.3 & 60.3 & 34.0 & 43.4 & 41.3 \\
        hi & 65.5 & 56.9 & 58.3 & 35.4 & 63.1 & 49.0 & 44.6 \\
        vi & 66.8 & 54.4 & 57.1 & 39.7 & 49.3 & 68.3 & 52.4 \\
        zh & 63.2 & 55.1 & 56.6 & 39.8 & 43.3 & 49.6 & 66.1 \\ \bottomrule
      \end{tabular}
    \end{center}
    \caption{MLQA full results of \mlukeW{}\ba{}}
    \label{table:mlqa-mlukeW-base}
  \end{minipage}
\end{table}

\begin{table}[h!]
  \small
  \center
  \begin{minipage}[t]{.45\textwidth}
    \begin{center}
      \begin{tabular}{cccccccc} \toprule
        c/q & en & es & de & ar & hi & vi & zh \\ \midrule
        en & 80.8 & 71.3 & 69.9 & 55.9 & 61.9 & 62.8 & 62.1 \\
        es & 70.6 & 69.9 & 66.4 & 52.6 & 53.7 & 57.6 & 58.0 \\
        de & 65.2 & 61.2 & 65.4 & 47.2 & 49.3 & 51.8 & 51.7 \\
        ar & 61.1 & 54.6 & 56.9 & 60.7 & 39.5 & 47.0 & 44.8 \\
        hi & 65.1 & 58.4 & 59.2 & 38.3 & 63.7 & 50.5 & 46.2 \\
        vi & 66.7 & 56.5 & 59.5 & 44.3 & 51.1 & 68.4 & 54.2 \\
        zh & 62.7 & 56.3 & 56.2 & 41.1 & 44.3 & 51.7 & 66.2 \\ \bottomrule
      \end{tabular}
    \end{center}
    \caption{MLQA full results of \mlukeE{}\ba{}}
    \label{table:mlqa-mlukeE-base}
  \end{minipage}
  \hspace{5mm}
  \begin{minipage}[t]{.45\textwidth}
    \begin{center}
  \begin{tabular}{cccccccc} \toprule
    c/q & en & es & de & ar & hi & vi & zh \\ \midrule
    en & 83.9 & 79.6 & 79.0 & 62.0 & 70.6 & 70.5 & 69.5 \\
    es & 75.2 & 74.7 & 73.0 & 60.3 & 63.4 & 66.6 & 65.9 \\
    de & 69.4 & 69.0 & 69.9 & 58.9 & 59.7 & 62.0 & 60.6 \\
    ar & 67.0 & 63.6 & 66.2 & 64.9 & 54.5 & 58.9 & 57.7 \\
    hi & 72.1 & 67.3 & 67.2 & 56.1 & 69.9 & 61.0 & 62.1 \\
    vi & 73.5 & 69.6 & 70.7 & 57.1 & 63.0 & 73.3 & 64.5 \\
    zh & 69.1 & 64.0 & 65.7 & 53.4 & 58.2 & 62.7 & 70.3 \\
    \bottomrule
  \end{tabular}
  \end{center}
  \caption{MLQA full results of \xlmr{}\la{}}.
  \label{table:mlqa-XLM-R-large}
  \end{minipage}
\end{table}

\begin{table}[h!]
  \small
  \center
  \begin{minipage}[t]{.45\textwidth}
    \begin{center}
      \begin{tabular}{cccccccc} \toprule
    c/q & en & es & de & ar & hi & vi & zh \\ \midrule
    en & 84.0 & 80.1 & 79.9 & 71.5 & 74.2 & 72.8 & 72.8 \\
    es & 74.6 & 74.3 & 74.6 & 65.5 & 64.3 & 66.0 & 66.0 \\
    de & 70.1 & 69.5 & 70.3 & 63.9 & 60.8 & 61.7 & 62.6 \\
    ar & 67.9 & 65.0 & 67.9 & 66.2 & 58.6 & 60.2 & 58.7 \\
    hi & 72.9 & 69.7 & 70.3 & 60.8 & 70.2 & 63.1 & 62.6 \\
    vi & 73.9 & 69.5 & 72.2 & 65.5 & 64.9 & 74.2 & 67.3 \\
    zh & 69.6 & 66.5 & 68.5 & 61.5 & 58.3 & 64.5 & 69.7 \\
    \bottomrule
      \end{tabular}
    \end{center}
    \caption{MLQA full results of \mlukeW{}\la{}}
    \label{table:mlqa-mlukeW-large}
  \end{minipage}
  \hspace{5mm}
  \begin{minipage}[t]{.45\textwidth}
    \begin{center}
      \begin{tabular}{cccccccc} \toprule
    c/q & en & es & de & ar & hi & vi & zh \\ \midrule
    en & 84.1 & 80.5 & 80.2 & 70.0 & 75.0 & 75.0 & 73.5 \\
    es & 75.2 & 74.5 & 74.8 & 62.4 & 65.3 & 67.6 & 66.5 \\
    de & 71.1 & 70.2 & 70.5 & 62.2 & 61.0 & 63.5 & 62.3 \\
    ar & 68.4 & 65.6 & 68.4 & 66.2 & 57.7 & 62.3 & 58.0 \\
    hi & 72.9 & 70.9 & 71.6 & 59.1 & 71.4 & 65.6 & 62.1 \\
    vi & 74.7 & 71.0 & 73.1 & 61.7 & 64.7 & 74.3 & 66.8 \\
    zh & 70.1 & 66.1 & 68.8 & 59.2 & 60.9 & 66.3 & 70.5 \\
    \bottomrule
      \end{tabular}
    \end{center}
    \caption{MLQA full results of \mlukeE{}\la{}}
    \label{table:mlqa-mlukeE-large}
  \end{minipage}
\end{table}

\newpage

\section{Full Results of mLAMA}
\label{sec:full_mlama}

\Table{table:mlama} shows the results from the setting where the entity candidates not in the \mluke{}'s entity vocabulary are excluded.
Here we provide in \Table{table:full_mlama} the results with the full candidate set provided in the dataset for ease of comparison with other literature. 
When the candidate entity is not found in the \mluke{}'s entity vocabulary, the log-probability from the word \mask{} tokens are used instead.

\begin{table}[h]
  
  \begin{tabular}{l|cccccccc} \toprule
         &   ar & bn & de & el & en & es & fi & fr  \\ \midrule
\mbert{} & 15.1 & 12.7 & 28.6 & 19.4 & 34.8 & 30.2 & 19.2 & 27.1  \\
\xlmr{}\ba & 14.9 & 7.5 & 18.4 & 12.7 & 24.2 & 18.5 & 14.5 & 16.1 \\
\extraTraining & 20.7 & 14.0 & 29.3 & 18.2 & 31.6 & 26.4 & 19.2 & 25.0 \\
\mlukeW{}\ba & 21.3 & 12.9 & 25.7 & 17.5 & 27.1 & 23.3 & 15.9 & 23.0 \\
\mlukeEwithY{} & 25.6 & 21.6 & 32.9 & 25.2 & 34.9 & 28.5 & 24.7 & 27.7 \\
\mlukeEwithXY{} & {\bf 37.3} & {\bf 32.3} & {\bf 43.7} & {\bf 34.4} & {\bf 43.2} & {\bf 36.4} & {\bf 35.3} & {\bf 34.2}  \\
  \end{tabular}

  \begin{tabular}{l|ccccccccc} \toprule
                & id & ja & ko & pl & pt & ru & vi & zh & avg. \\ \midrule
\mbert{} &  37.4 & 14.2 & 17.8 & 21.9 & 32.0 & 17.4 & 36.5 & 24.2 & 24.3 \\
\xlmr{}\ba  & 24.6 & 11.4 & 10.9 & 16.6 & 22.2 & 12.6 & 23.0 & 15.5 & 16.5   \\
\extraTraining & 38.2 & 19.1 & 21.4 & 20.5 & 29.6 & 20.6 & 33.8 & 28.1 & 24.7   \\
\mlukeW{}\ba & 36.6 & 18.0 & 17.9 & 20.2 & 29.4 & 19.6 & 31.0 & 26.9 & 22.9  \\
\mlukeEwithY{} & 35.3 & 27.2 & 26.3 & 25.7 & 34.7 & 23.8 & 39.1 & 29.5 & 28.9   \\
\mlukeEwithXY{} & {\bf 47.6} & {\bf 37.7} & {\bf 41.6} & {\bf 37.7} & {\bf 44.8} & {\bf 31.4} & {\bf 50.1} & {\bf 41.6} & {\bf 39.3}  \\ \bottomrule
  \end{tabular}

  \caption{The average of Top-1 accuracies from 16 languages from the mLAMA dataset.}
  \label{table:full_mlama}
\end{table}


\end{document}